\documentclass[10pt,twocolumn,letterpaper]{article}

\usepackage{cvpr}              
\usepackage{listings}

\usepackage[dvipsnames]{xcolor}

\definecolor{cvprblue}{rgb}{0.21,0.49,0.74}
\usepackage[pagebackref,breaklinks,colorlinks,citecolor=cvprblue]{hyperref}
\usepackage{graphicx} 
\usepackage{grffile}

\def\paperID{22} 
\def\confName{CVPR}
\def\confYear{2024}

\title{SINGLE-IMAGE COHERENT RECONSTRUCTION OF OBJECTS AND HUMANS}

\author{Sarthak Batra\\
University of Surrey\\
{\tt\small s.batra@surrey.ac.uk}\\
\\
Partha P. Chakrabarti \\
Indian Institute of Technology Kharagpur\\
{\tt\small ppchak@cse.iitkgp.ac.in}
\and
Simon Hadfield\\
University of Surrey\\
{\tt\small s.hadfield@surrey.ac.uk}\\
\\
Armin Mustafa\\
University of Surrey\\
{\tt\small armin.mustafa@surrey.ac.uk}\\
}

\begin{document}
\maketitle
\begin{abstract}
Existing methods for reconstructing objects and humans from a monocular image suffer from severe mesh collisions and performance limitations for interacting occluding objects. This paper introduces a method to obtain a globally consistent 3D reconstruction of interacting objects and people from a single image.
Our contributions include: 1) an optimization framework, featuring a collision loss, tailored to handle human-object and human-human interactions, ensuring spatially coherent scene reconstruction; and 2) a novel technique to robustly estimate 6 degrees of freedom (DOF) poses, specifically for heavily occluded objects, exploiting image inpainting. Notably, our proposed method operates effectively on images from real-world scenarios, without necessitating scene or object-level 3D supervision. Extensive qualitative and quantitative evaluation against existing methods demonstrates a significant reduction in collisions in the final reconstructions of scenes with multiple interacting humans and objects and a more coherent scene reconstruction.
\end{abstract}    
\vspace{-0.5cm}
\section{Introduction}
\label{sec:intro}
Existing methods for human and object reconstructions are either limited to single objects and humans or give limited performance for complex images with multiple people and objects \cite{kanazawa2018end,kolotouros2019learning,choy20163d,girdhar2016learning,hassan2019resolving,savva2016pigraphs}. 
These methods estimate the 3D poses of humans and objects independently and do not take into account the human-human interactions \cite{zhang2020perceiving} and even if they do they generally follow a supervised approach \cite{jiang2020coherent}. This leads to large collisions between the meshes with incoherent reconstructions. We consider the full scene holistically and exploit information from the human-human and human-object interactions for spatially coherent and more complete 3D reconstruction of in-the-wild images.

\begin{figure*}[t]
\includegraphics[scale = 0.65]{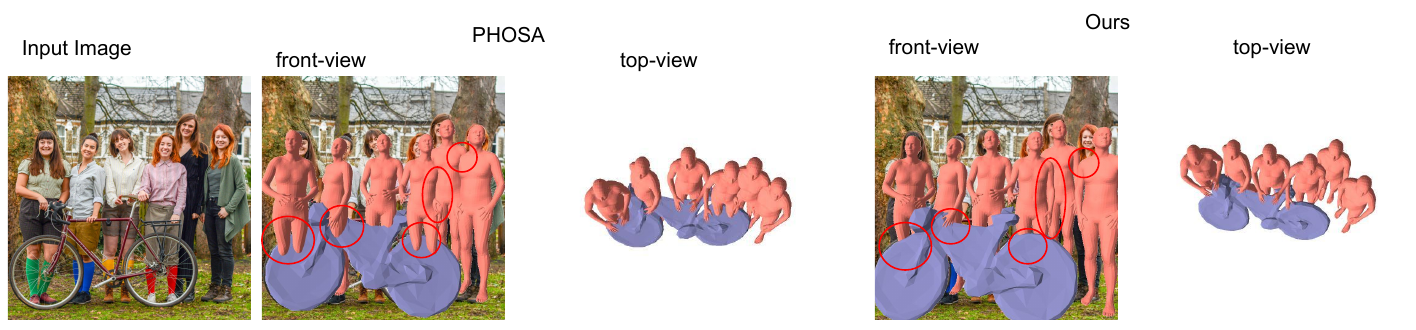}
\centering
\label{fig:motivation}
\vspace{-0.35cm}
\caption{Comparison of the proposed method (right) reconstruction with PHOSA(middle). The proposed method gives a more coherent reconstruction with correct spatial arrangement by reasoning about human-human and human-object interaction}
\vspace{-0.6cm}
\end{figure*}

PHOSA \cite{zhang2020perceiving} pioneered the field and proposed the first method that reconstructs humans interacting with objects for in-the-wild images. However, PHOSA does not explicitly model human-human interactions and gives erroneous reconstructions when objects are heavily occluded which leads to reconstructions with incorrect depth ordering and mesh collisions. Multi-human model-free reconstruction from a single image was proposed in \cite{mustafa2021multi}, however, this method does not deal with interacting humans. 
Other methods \cite{sun2021monocular, sun2022putting} for multi-human reconstructions generate reconstructions with severe mesh collisions because they reconstruct each person independently. To address these challenges, in this paper, we have proposed an optimization-based framework for the spatially coherent reconstruction of scenes with multiple interacting people and heavily occluded objects. 
The method first reconstructs humans \cite{joo2021exemplar} and objects \cite{kato2018neural} in the image independently. The initial poses of people in the scene are then optimized to resolve any ambiguities that arise from this independent composition using a collision loss, depth ordering loss, and interaction information.
To deal with heavily occluded objects, a novel 6 DOF pose estimation is proposed that uses inpainting to refine the segmentation mask of the occluded object for significantly improved pose estimation. Finally, we propose a global objective function that scores 3D object layouts, orientations, collision, and shape exemplars. Gradient-based solvers are used to obtain globally optimized poses for humans and objects. Our contributions are:
\begin{itemize}[topsep=0pt,partopsep=0pt,itemsep=0pt,parsep=0pt] 
	\item A method for generating a cohesive scene reconstruction from a single image by capturing interactions among humans and between humans and objects within the scene, all without relying on any explicit 3D supervision.
	\item A collision loss in an optimization framework to robustly estimate 6 DOF poses of multiple people and objects in crowded images.
	\item An inpainting-based method to improve the segmentation mask of heavily occluded objects that greatly boosts the precision of 6 DOF object position estimations.
	\item Extensive evaluation of the proposed method on complex images with multiple interacting humans and objects from the COCO-2017 dataset \cite{lin2014microsoft} against the state-of-the-art demonstrate the effectiveness of our approach.
\end{itemize}


\section{Related Work}
\label{sec:related work}

\textbf{3D humans from a single image: }Reconstructing 3D human models from images is often achieved through various methodologies. One widely used approach involves fitting parametric models like SMPL\cite{loper2023smpl} to input images \cite{alldieck2018detailed,alldieck2018video,bogo2016keep,corona2022learned,kolotouros2019learning,pavlakos2019expressive,pons2011model}. Alternatively, learning-based techniques directly predict model parameters such as pose and shape\cite{alldieck2019learning,guler2019holopose,jiang2020coherent,kanazawa2018end,kocabas2020vibe,omran2018neural,pavlakos2018learning}. \cite{loper2015smpl,pavlakos2019expressive,zhou2010parametric} use statistical body models and a large number of 3D scans to recover 3D humans from a single image. \cite{bogo2016keep} use 2D poses, \cite{tung2017self} uses 2D body joint heatmaps and \cite{kolotouros2019convolutional} uses GraphCNN to estimate SMPL model \cite{loper2015smpl}. However, these methods only estimate the 3D of a single person in the scene. Methods like \cite{zanfir2018monocular,zanfir2018deep,jiang2020coherent,mustafa2021multi} recover the 3D poses and shapes of multiple people focus on resolving ambiguities that arise due to incorrect depth ordering and collisions between people. However, these methods cannot handle large occlusions.  Recent advancements in whole-body mesh recovery from images have shifted the focus from solely on regressing body parameters to also accurately estimating hand and face parameters. An example is FrankMocap \cite{rong2021frankmocap}, which employs a modular design. This approach initially runs independent 3D pose regression methods for the face, hands, and body and integrates their outputs through a dedicated module. A one-stage pipeline named OSX \cite{lin2023one} has been introduced for 3D whole-body mesh recovery, surpassing existing multi-stage models in accuracy. It introduces a component-aware transformer (CAT) comprising a global body encoder and a local face/hand decoder. KBody\cite{zioulis2023kbody} represents a methodology for fitting a low-dimensional body model to an image, employing a predict-and-optimize approach. A distinctive feature of KBody is the introduction of virtual joints, enhancing correspondence quality and disentangling the optimization process between pose and shape parameters.

\begin{figure*}[th]
\includegraphics[scale = 0.65]{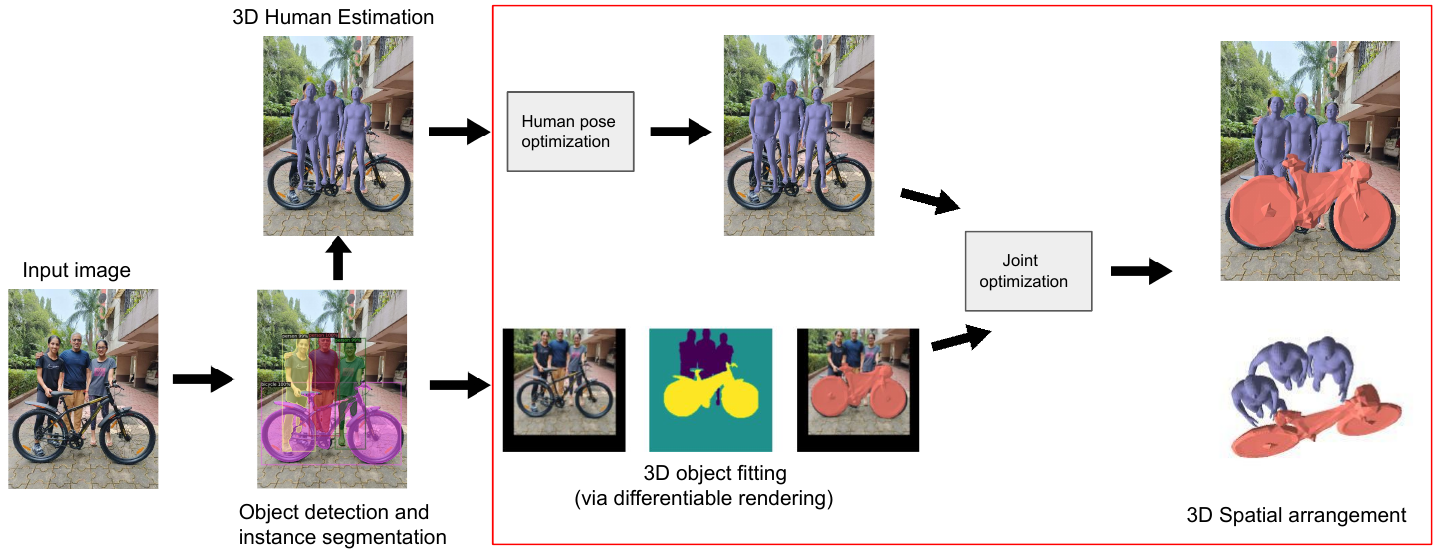}
\centering
\caption{Overview of the proposed method to generate spatially coherent reconstruction from a single image. The steps in red box are novel. The reconstruction before human pose optimization exhibits notable mesh collisions. After human pose optimization, reduced mesh collisions are seen while
maintaining relative coherence between humans.}
\vspace{-0.6cm}
\label{fig:2}
\end{figure*}
\noindent
\textbf{3D objects from a single image: } Single-view 3D reconstruction is a complex task, as it necessitates incorporating reliable geometric priors derived from our 3D world. However, these priors often lack in diverse real-world scenarios \cite{ikeuchi1981numerical,horn1970shape,pentland1984local, pentland1988shape}. Given their robustness and accessibility, learning-based methods have emerged. Deep learning approaches can be categorized based on the employed 3D representations, encompassing voxel-based frameworks \cite{choy20163d,mescheder2019occupancy,richter2018matryoshka,popov2020corenet,wu2016learning}, point cloud-based methods \cite{fan2017point,han2020drwr,chen2021unsupervised}, mesh-based techniques \cite{gkioxari2019mesh,liu2019soft,li2020self,kanazawa2018learning}, and implicit function-based approaches \cite{chen2019learning,saito2019pifu,saito2020pifuhd}. The majority of current single-view 3D mesh reconstruction methods employ an encoder-decoder framework. Here, the encoder discerns perceptual features from the input image, while the decoder distorts a template to conform to the desired 3D shape. The pioneering work by \cite{wang2018pixel2mesh} introduced deep learning networks to this task. They employed the VGG network \cite{simonyan2014very} as the encoder and utilized a graph convolutional network (GCN) as the decoder. \cite{groueix2018papier} introduced a method wherein a 3D shape is represented as a collection of parametric surface elements, allowing for a flexible representation of shapes with arbitrary topology. \cite{pan2019deep} addressed topology changes by proposing a topology modification network that adaptively deletes faces. These methods are trained and evaluated on identical object categories.

Recent research has also devised techniques for 3D reconstruction from image collections without explicit 3D supervision. This has been achieved by employing differentiable rendering to supervise the learning process. For instance, \cite{kanazawa2018learning} proposed a method that reconstructs the underlying shape by learning deformations on top of a category-specific mean shape. \cite{lin2020sdf} developed a differentiable rendering formulation to learn signed distance functions as implicit 3D shape representations, overcoming topological restrictions. \cite{duggal2022topologically} learned both deformation and implicit 3D shape representations, facilitating reconstruction in category-specific canonical space. \cite{vasudev2022pre} extended category-specific models into cross-category models through distillation. \cite{kulkarni20193d} used GNN trained on a synthetic dataset without any humans to deduce an object's layout. 

\noindent
\textbf{3D human-to-object interaction: }Modeling 3D human-object interactions poses significant challenges. Recent studies have demonstrated remarkable success in capturing hand-object interactions from various perspectives, including 3D \cite{brahmbhatt2019contactgrasp,taheri2020grab,zhou2022toch}, 2.5D \cite{brahmbhatt2019contactdb,brahmbhatt2020contactpose}, and images \cite{corona2020ganhand,ehsani2020use,hasson2019learning,karunratanakul2020grasping}. However, these achievements are limited to hand-object interactions and do not extend to predicting the full body. The complexity increases when considering full-body interactions, with works like PROX \cite{hassan2019resolving} successfully reconstructing \cite{hassan2019resolving,weng2021holistic} or synthesizing \cite{hassan2021populating,zhang2020place,zhang2022couch} 3D humans to adhere to 3D scene constraints. Other approaches focus on capturing interactions from multiple views \cite{bhatnagar2022behave,jiang2022neuralhofusion,sun2021neural} or reconstructing 3D scenes based on human-scene interactions \cite{yi2022human}. More recently, efforts have extended to model human-human interactions \cite{fieraru2020three} and self-contacts \cite{fieraru2021learning,muller2021self}. \cite{savva2016pigraphs} used information from the RGBD videos of individuals interacting with interiors to train a model that understands how people interact with their surroundings. Access to 3D scenes gives scene constraints that enhance the perception of 3D human poses \cite{yamamoto2000scene,rosenhahn2008markerless,kjellstrom2010tracking}. \cite{hassan2019resolving} uses an optimization-based method to enhance 3D human posture estimates conditioned on a particular 3D scene obtained from RGBD sensors. Another recent method, \cite{rosinol20203d}, creates a 3D scene graph of people and objects for indoor data. \cite{chen2019holistic++} represents the optimal configuration of the 3D scene, in the form of a parse graph that encodes the object, human pose, and scene layout from a single image. In our work, we overcome the limitations of existing methods by handling not only on human-object interactions but also capturing human-human interactions and propose a method that deals with major occlusions to significantly improved scene reconstruction.

\section{Methodology}
\label{sec:methodology}

The proposed method takes a single RGB image as input and gives a spatially coherent reconstruction of interacting humans and objects in the scene, an overview is shown in Figure \ref{fig:2}. We exploit human-human and human-object interactions to spatially arrange all objects in a common 3D coordinate system.
First, objects and humans are detected, followed by SMPL-based per-person reconstruction (Sec. \ref{sec:3.1}), which gives incorrect spatial reconstructions with collisions between meshes. 
The human 3D locations/poses are translated into world coordinates and refined through a human-human spatial arrangement optimization using a collision loss (Sec. \ref{sec:3.2}). To correctly estimate the 3D object pose (6-DoF translation and orientation) a differentiable renderer is used that fits 3D mesh object models to the predicted 2D segmentation masks \cite{kirillov2020pointrend}. We correct the occluded object mask using image inpainting (in Sec. \ref{sec:3.3}) unlike PHOSA \cite{zhang2020perceiving} which uses an occluded object mask.
Lastly, we perform joint optimization that takes into account both human-human and human-object interactions for a globally consistent output. Our framework produces plausible reconstructions that capture realistic human-human and human-object interactions.

\begin{figure*}[t]
\includegraphics[scale = 0.75]{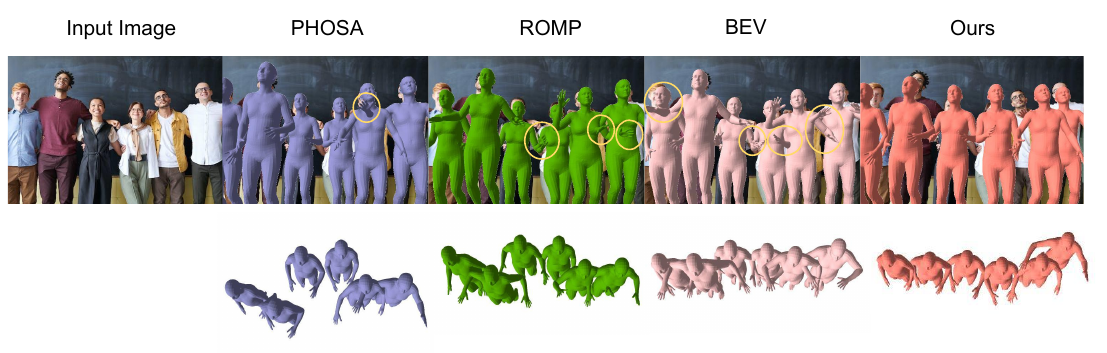}
\centering
\vspace{-0.5cm}
\caption{The proposed approach gives spatially coherent reconstructions with a significant reduction in mesh collisions compared to PHOSA \cite{zhang2020perceiving}, ROMP\cite{sun2021monocular}, and BEV\cite{sun2022putting}. Significant collision are shown in highlighted circles.} 
\vspace{-0.5cm}
\label{fig:meshCollisions}
\end{figure*}
\subsection{Estimating 3D Humans} \label{sec:3.1}
Using \cite{joo2021exemplar}, we estimate the 3D shape and pose parameters of SMPL \cite{loper2015smpl} given a bounding box for a human \cite{mask2017cnn}. The 3D human is parameterized by pose $\theta \in \mathbb{R}^{72}$, shape $\beta \in \mathbb{R}^{10}$, and a weak-perspective camera $\gamma$ = [$\sigma$, $t_{x}, t_{y}$] $\in \mathbb{R}^{3}$. To position the humans in the 3D space, $\gamma$ is converted to the perspective camera projection by assuming a fixed focal length f for all images, and the distance of the person is determined by the reciprocal of the camera scale parameter $\sigma$. Thus, the 3D vertices of the SMPL model for the $i^{th}$ human are represented as: $M_{i} = J(\theta_{i},\beta_{i}) + [t_{x}, t_{y}, \frac{f}{\sigma^{i}}]$,
where $J$ is the differentiable SMPL mapping from pose and shape to a human mesh and $t_{h}^{i}=[t_{x}, t_{y}, \frac{f}{\sigma^{i}}]$ is the translation of $i^{th}$ human. The person's height and size are regulated by the SMPL shape parameter $\beta$. We define scale parameter($s^{i}$) for each human similar to PHOSA and the final vertices are given by $\bar{M}_{i} = s^{i}M_{i}$.

\subsection{Human Pose Optimisation} \label{sec:3.2}
Independently analyzing human 3D poses results in inconsistent 3D scene configurations. Human-human interactions offer useful information to determine the relative spatial arrangement and not considering this leads to ambiguities like mesh penetration and incorrect depth ordering. We propose an optimization framework that incorporates human-human interactions. We first identify interacting humans in the image and then optimize the pose through an objective function to correctly adjust their spatial arrangements.

\noindent
\textbf{Identifying interacting humans - } Our hypothesis posits that human interactions are contingent upon physical proximity in world coordinates. Hence we find the interacting humans by identifying the overlap of 3D bounding boxes(More details regarding bounding box overlap criteria can be found in the supplementary material).

\noindent
\textbf{Objective function to optimize 3D spatial arrangement} Our objective includes collision ($L_{H-collision}$), interaction ($L_{H-interaction}$), and depth ordering loss ($L_{H-depth}$) terms to constraint the pose for interacting humans:
\vspace{-0.4cm}
\begin{multline}\label{eq:2}
        L_{HHI-Loss} = \lambda_{1} L_{H-collision} + \lambda_{2} L_{H-depth} \\ + \lambda_{3} L_{H-interaction}
\end{multline}

We optimize (\ref{eq:2}) using a gradient-based optimizer \cite{kingma2014adam} w.r.t. translation $\textbf{t}^{i}$ $\in \mathbb{R}^3$ and scale parameter $s^{i}$ and the Rotation $\textbf{R}^{i}$ $\in SO3$ for the $i^{th}$ human instance . The human translations are initialized from Sec \ref{sec:3.1}. The terms in the objective function are defined below:

\noindent
\textbf{Collision Loss ($L_{H-collision}$) - } To overcome the problem of mesh collisions, as seen in existing methods in \cref{fig:meshCollisions}, we introduce a collision loss  $L_{H-collision}$ that penalizes interpenetrations in the reconstructed people. Let $\phi$ be a modified Signed Distance Field (SDF) for the scene that is defined as follows: $\phi(x,y,z) = - min(SDF(x,y,z),0)$
where $\phi$ is positive for points inside the human and is proportional to the distance from the surface, and is 0 outside of the human. Typically $\phi$ is defined on a voxel grid of dimensions $N_{p} * N_{p} * N_{p}$. While it's definitely possible to generate a single voxelized representation for the entire scene, we often find ourselves requiring an extensive fine-grained voxel grid. Depending on the scene's extent, this can pose processing challenges due to memory and computational limitations. To overcome this a separate $\phi_{i}$ function is computed for each person by calculating a tight box around the person and voxelizing it instead of the whole scene to reduce computational complexity \cite{jiang2020coherent}. The collision penalty of person $j$ for colliding with person $i$ is defined as follows:
   $ P_{ij} = \sum_{v \in {M}_{j}} \tilde\phi_{i}(v) $, where $\tilde\phi_{i}(v)$ samples the $\phi_{i}$ value for each 3D vertex $v$ in a differentiable way from the 3D grid using trilinear interpolation.
If there is a collision between person $i$ and a person $j$, $P_{ij}$ will be a positive value and decreases as the separation between them increases. If there is no overlap between person $i$ and $j$, $P_{ij}$ will be zero.
Let the translation of person $i$ and person $j$ be $T_{i}$ and $T_{j}$ respectively. Then the collision loss between them is defined as:
\vspace{-0.0cm}
\begin{equation} \label{eq:4}
  L_{ij} =
    \begin{cases}
      \frac{ P_{ij}}{\exp({||T_{i}-T_{j}} + \delta||_{2})} & \text{$T_{i} = T_{j}$}\\
      \frac{ P_{ij}}{\exp({||T_{i}-T_{j}}||_{2})} & \text{$T_{i} \neq T_{j}$}
    \end{cases}   
    \vspace{-0.2cm}
\end{equation}

When the translation values are the same (in case of maximum overlap) we use an extra term $\delta$ ($0<\delta<1$) to ensure non-zero gradients are not very large to avoid any instabilities during optimization. The final collision loss for a scene with N people is defined as follows:
\vspace{-0.1cm}
\begin{equation} \label{eq:6}
  L_{H-collision} = \sum_{j=1}^{N}  \Bigl (\sum_{i=1 i\neq j}^{N} L_{ij} \Bigl)  
  \vspace{-0.2cm}
\end{equation}

\noindent
\textbf{Interaction Loss ($L_{H-interaction}$) - }This is an instance-level to pull the interacting people close together, similar to \cite{zhang2020perceiving}: $L_{H-interaction} = \sum_{h_{i},h_{j} \in H} \mu(h_{i},h_{j})||C(h_{i}) - C(h_{j})||_{2}$, 
where $\mu(h_{i},h_{j})$ identifies whether human $h_{i}$ and $h_{j}$ are interacting according to the 3D bounding box overlap criteria. $C(h_{i})$ and $C(h_{j})$ give the centroid for human $i$ and human $j$ respectively.

\begin{figure*}[h!]
\includegraphics[scale = 0.7]{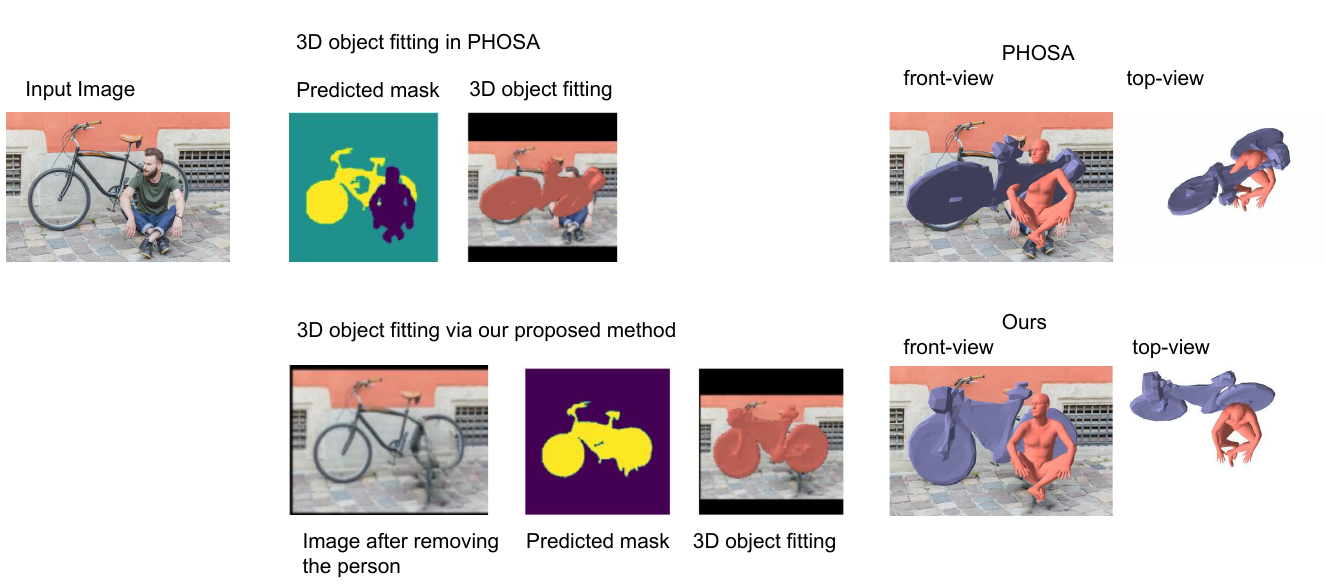}
\centering
\vspace{-0.2cm}
\caption{Comparison of the segmentation masks and reconstruction with PHOSA. The segmentation mask of the bicycle is occluded resulting in erroneous reconstruction in PHOSA. The proposed method uses image inpainting to remove the occlusion to generate a better segmentation mask, which leads to a more complete reconstruction.}
\vspace{-0.6cm}
\label{fig:3Dobjects}
\end{figure*}

\noindent
\textbf{Depth-Ordering Loss ($L_{H-depth}$) - } This helps to achieve more accurate depth ordering, as in \cite{jiang2020coherent}. The loss is defined as: $L_{depth} = \sum_{p \in S} \log(1 + \exp(D_{y(p)}(p) - D_{\Bar{y}(p)}(p) ))$, where $S = \{ p  \in  I: y(p) > 0, \Bar{y}(p) > 0,y(p) \neq \Bar{y}(p)\}$ is the pixels in the image I with incorrect depth ordering in the ground truth segmentation, the person index at pixel position $p$ is represented by $y(p)$, and the predicted person index in the rendered 3D meshes is $\Bar{y}(p)$ and $y(p) \neq \Bar{y}(p)$. $D_{y(p)}(p)$ and $D_{\Bar{y}(p)}(p)$ represent the pixel depths.

\subsection{3D Object Pose Estimation} \label{sec:3.3}
After estimating the shape and pose of humans, the next step is to estimate the same for the objects. To estimate the 3D location $t \in \mathbb{R}^3$ and 3D orientation $R \in SO(3)$ of the objects. For each object category, exemplar mesh models are pre-selected. The mesh models are sourced from \cite{Free3D:2013,kundu20183d}. The vertices of $j^{th}$ object are: $V_{o}^{j} = s^{j}(R^{j}O(c^{j},k^{j}) + t^{j})$, where $c^{j}$ is the object category from MaskRCNN \cite{mask2017cnn}, and $O(c^{j},k^{j})$ determines the $k^{j}-th$ exemplar mesh for category $c^{j}$. The optimization framework chooses the exemplar that minimizes re-projection error to determine $k^{j}$ automatically and $s^{j}$ is the scale parameter for $j^{th}$ object.

Our first objective is to estimate the 6 DOF pose of each object independently. It is difficult to estimate 3D object pose in the wild as there are: (1) no parametric 3D models for objects; (2) no images of objects in the wild with 2D/3D pose annotations; and (3) occlusions in cluttered scenes with humans. We address these challenges by proposing an optimization-based approach that uses a differentiable renderer \cite{kato2018neural} to fit the 3D object to instance masks from \cite{kirillov2020pointrend} in a manner that is robust to minor/major occlusions. 

As defined in \cite{zhang2020perceiving} we calculate a pixel-wise L2 loss over rendered silhouettes S versus predicted masks M but the quality of the predicted mask M is impacted by occlusions as seen in \cite{zhang2020perceiving}, which results in a poorly estimated 6 DOF pose.
To address problems due to occlusions, we propose a novel method that improves the masks as shown in \cref{fig:3Dobjects}.

Given an image $I$, a total number of objects $N$, and bounding boxes for rigid $B_{r}$ and non-rigid $B_{nr}$ objects, along with their masks - $M_{r}$ for rigid and $M_{nr}$ for non-rigid objects.
Each $i^{th}$ object can be occluded by maximum $N-1$ objects. To identify occluding objects we calculate the Intersection over Union(IOU) between all pairs of bounding boxes. Objects with $IOU>0.3$ (Our selection of this threshold stems from our empirical observations, wherein we found that objects with $IOU>0.3$ led to noticeable improvements in reconstruction quality. Conversely, when IOU was less than 0.3, the reconstruction results obtained using our method closely resembled those produced by PHOSA \cite{zhang2020perceiving}, more details in supplementary) are occluding objects $M$ for each object. 
Occluding objects can be removed in numerous ways, for e.g remove only one object at a time. The total possible combinations, in this case, are $\binom{M}{1}$, or you remove a pair of objects at a time and the total possible combinations, in this case, are $\binom{M}{2}$ and so on. The total number of all possible combinations can be described as $\binom{M}{0}+\binom{M}{1}+\binom{M}{2}+.....\binom{M}{M} = 2^{M}$. To remove $j$ occluding objects where $j \le M$ we need a single mask $M_{occ-mask}$ that is a combination of the $j$ masks, so $M_{occ-mask} = M_{1} + M_{2} + .... + M_{j}$. 
Now we use the image-inpainting approach proposed by \cite{nazeri2019edgeconnect} to remove the occluding objects. We pass the current image $I$ and the mask $M_{occ-mask}$ to get a new image without occlusions and use this image to get the new segmentation masks and bounding boxes:
\vspace{-0.0cm}
\begin{equation} \nonumber
  I_{new} = EC(I, M_{occ-mask})
\end{equation}
\begin{equation}
  B_{r}^{new},B_{nr}^{new},M_{r}^{new},M_{nr}^{new} = OD(I_{new})
\end{equation}

where $EC$ is the image inpainting algorithm and $OD$ is the object detection algorithm.
Sometimes, the $i^{th}$ object in $I$ may not correspond to the same object in $I_{new}$. Let's say the index of the $i^{th}$ object in $I_{new}$ be $k$. We iterate over the list of new bounding boxes and calculate the $IOU$ of these boxes with $B_{r}[i]$ and, $k$ corresponds to the index of the bounding box for which $IOU$ is closest to 1. We use the mask $M_{r}^{new}[k]$ to determine object pose. Estimating a reliable pose also depends heavily on the boundary details. To incorporate this we augment the L2 mask loss with a modified version of the symmetric chamfer loss \cite{gavrila2000pedestrian}. Given a no-occlusion indicator $\eta$ (0 if pixel only corresponds to a mask of a different instance, else 1), the loss is:
\vspace{-0.0cm}
\begin{equation} \label{eq:12}
  L_{occ-sil} = \sum (\eta \circ S - M_{r}^{new}[k])^2 + \sum_{p \in E(\eta \circ S)}\underset{\Bar{p} \in E(M)}{min}||p-\Bar{p_{2}}||
  \vspace{-0.1cm}
\end{equation}

We generate masks $M_{occ-mask}$ for different values of $j$ and a 3D pose corresponding to that mask is chosen that results in a minimum value of $L_{occ-sil}$. The edge map of mask M is computed by E(M). To estimate the 3D object pose, we minimize the occlusion-aware silhouette loss:

\begin{equation}
  (R^{j}, t^{j})^{*} = \underset{R,t}{argmin}  (L_{occ-sil}(\Pi_{sil}(V_{o}^{j}), M_{r}^{new}[k]))
\end{equation}

where $\Pi_{sil}$ is the silhouette rendering of a 3D mesh model via a perspective camera with a fixed focal length f (Sec \ref{sec:3.1}) and $M_{j}$ is a 2D instance mask for the $j^{th}$ object. Instance masks are computed by PointRend \cite{kirillov2020pointrend}.

\subsection{Joint Optimization}
The joint optimization refines both the human and object poses estimated above, exploiting both human-human and human-object interactions through joint loss functions. 
Estimating 3D poses of people and objects in isolation from one another leads to inconsistent 3D scene reconstruction. Interactions between people and objects provide crucial clues for correct 3D spatial arrangement, which is done by identifying interacting objects and humans and proposing an objective function for refining human/object poses.

\noindent
\textbf{Identifying human-object interaction}. Our hypothesis posits that human-object interactions are contingent upon physical proximity in world coordinates. We use 3D bounding box overlap between the human and object to determine whether the object is interacting with a person.

\noindent
\textbf{Objective function to optimize 3D spatial arrangements}. 
We define a joint loss function that takes into account both human-human and human-object interactions. It is crucial to include both of them because if you simply optimize with regard to human-object interactions, it may result in erroneous relative positions between interacting people even if it would enhance the relative spatial arrangement between the interacting humans and objects.
\begin{equation} \label{eq:method}
     L_{joint-loss} = L_{HOI-Loss} + L_{HHI-Loss}
\end{equation}
where $L_{HHI-Loss}$ is same as Eq. \ref{eq:2} and
\begin{multline}\label{eq:14}
    L_{HOI-Loss} = \lambda_{1} L_{HO-collision} + \lambda_{2} L_{HO-depth} \\ + \lambda_{3} L_{HO-interaction} + \lambda_{4}L_{occ-sil}
\end{multline}

{Depth-Ordering Loss ($L_{HO-depth}$)} is same as Section \ref{sec:3.2}.
We optimize (\ref{eq:14}) using a gradient-based optimizer \cite{kingma2014adam} w.r.t. translation $t^{i} \in \mathbb{R}^3$ and intrinsic scale $s^{i} \in \mathbb{R}$  for the $i^{th}$ human and, rotation $R^{j} \in SO(3)$, translation $t^{j} \in \mathbb{R}^3$ and $s^{j} \in \mathbb{R}$ for the $j^{th}$ object instance jointly. The object poses are initialized from Sec. \ref{sec:3.3}. $L_{occ-sil}$ is the same as (\ref{eq:12}) except without the chamfer loss which didn’t help during joint optimization. 

\noindent
\textbf{Interaction loss ($L_{HO-interaction}$)}: This loss handles both coarse and fine interaction between humans and objects as in \cite{zhang2020perceiving}, defined as: $L_{HO-interaction} = L_{coarse-inter} + L_{fine-inter}$.\\
The coarse interaction loss is: $L_{coarse-inter} = \sum_{h \in H,o \in O} \mu(h,o)||C(h) - C(o)||_{2}$, where $\mu(h,o)$ identifies whether human $h$ and object $o$ are interacting according to the 3D bounding box overlap criteria. $C(h)$ and $C(o)$ give the centroid for human and the object respectively. 
To handle human interactions, the fine interaction loss is defined as: \\$L_{fine-inter} = \sum_{h \in H,o \in O} (\sum_{P_{h},P_{o} \in P(h,o)} \mu(P_{h},P_{o})||C(P_{h}) - C(P_{o})||_{2})$, where $P_{h}$ and $P_{o}$ are the regions of interaction between the humans and the object, respectively. $\mu(P_{h}, P_{o})$ is the overlap of the 3D bounding box between the interacting objects, recomputed at each iteration.

\noindent
\textbf{Collision Loss ($L_{HO-collision}$) - }The formulation of this loss is similar to the collision loss defined in Section \ref{sec:3.2}. The difference is that here we take into account the mesh collision between interacting humans and objects in contrast to interacting humans. Let $N_{h}$ represent the total number of humans and $N_{o}$ total number of objects, then the Loss function is defined as: $L_{HO-collision} = \sum_{j=1}^{N_o}  \Bigl (\sum_{i=1}^{N_h} L_{h_{i}o_{j}} + L_{o_{j}h_{i}} \Bigl) $, where $h_{i}$ represents the $i^{th}$ human and $o_{j}$ represents the $j^{th}$ object.

\section{Results and Evaluation}

We perform both quantitative and qualitative assessments of the performance of our technique on the COCO-2017 \cite{lin2014microsoft} dataset on images that include interactions of humans and objects against PHOSA \cite{zhang2020perceiving}, ROMP\cite{sun2021monocular}, and BEV\cite{sun2022putting}.

\subsection{Qualitative and Quantitative Analysis}\label{sec:eval}

\begin{figure} 
\includegraphics[scale = 0.4]{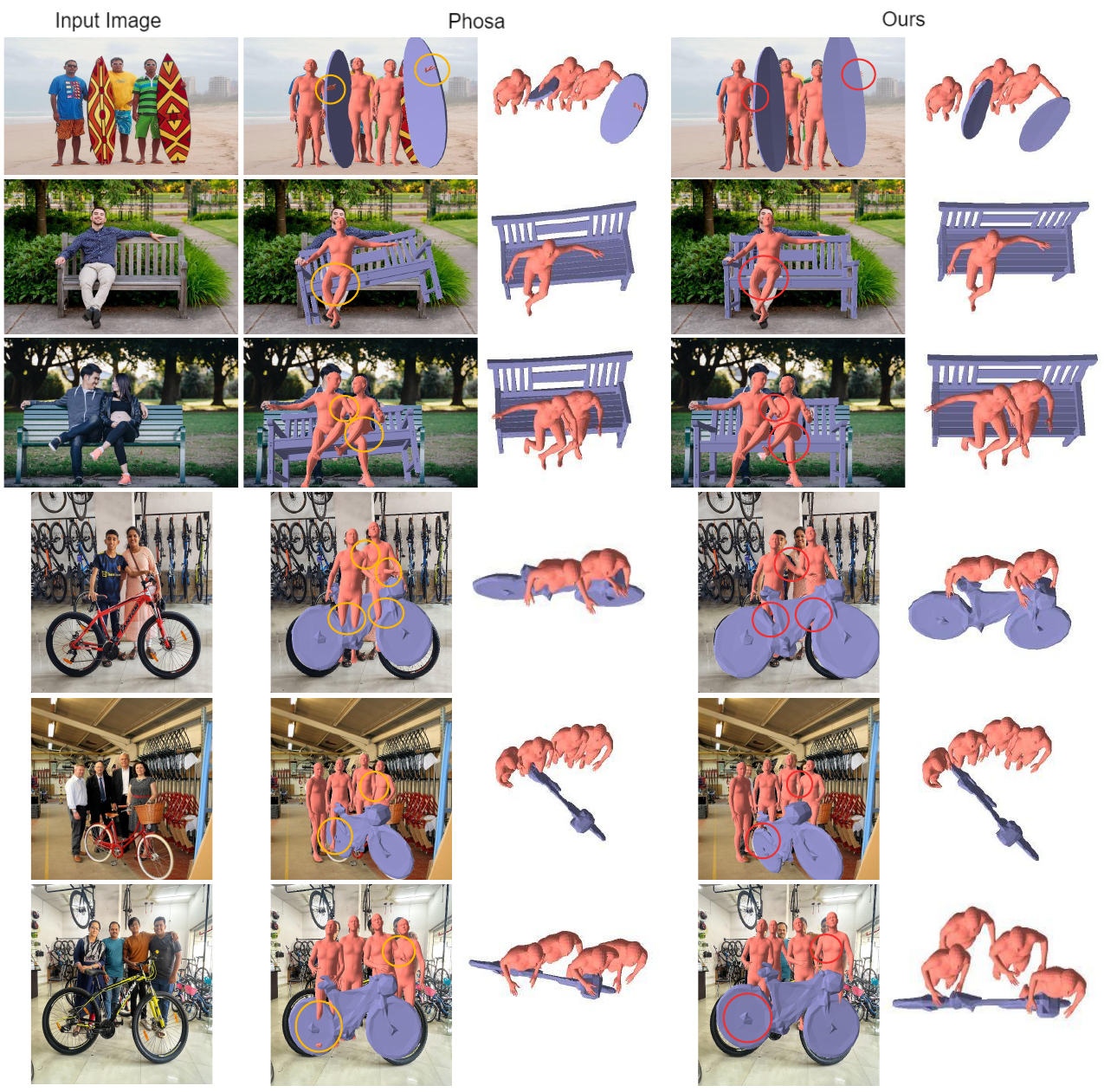}
\centering

\caption{Qualitative comparison on test images from COCO 2017 against PHOSA \cite{zhang2020perceiving} with human-object interactions. Our method gives better spatial reconstruction while substantially reducing collisions(the golden circles delineate regions characterized by noteworthy mesh collisions, while the red circles delineate areas showcasing enhancements in reconstructions). More qualitative results are shown in \ref{sec:4.3}}
\label{fig:5}
\end{figure}

\begin{figure}
\includegraphics[scale = 0.4]{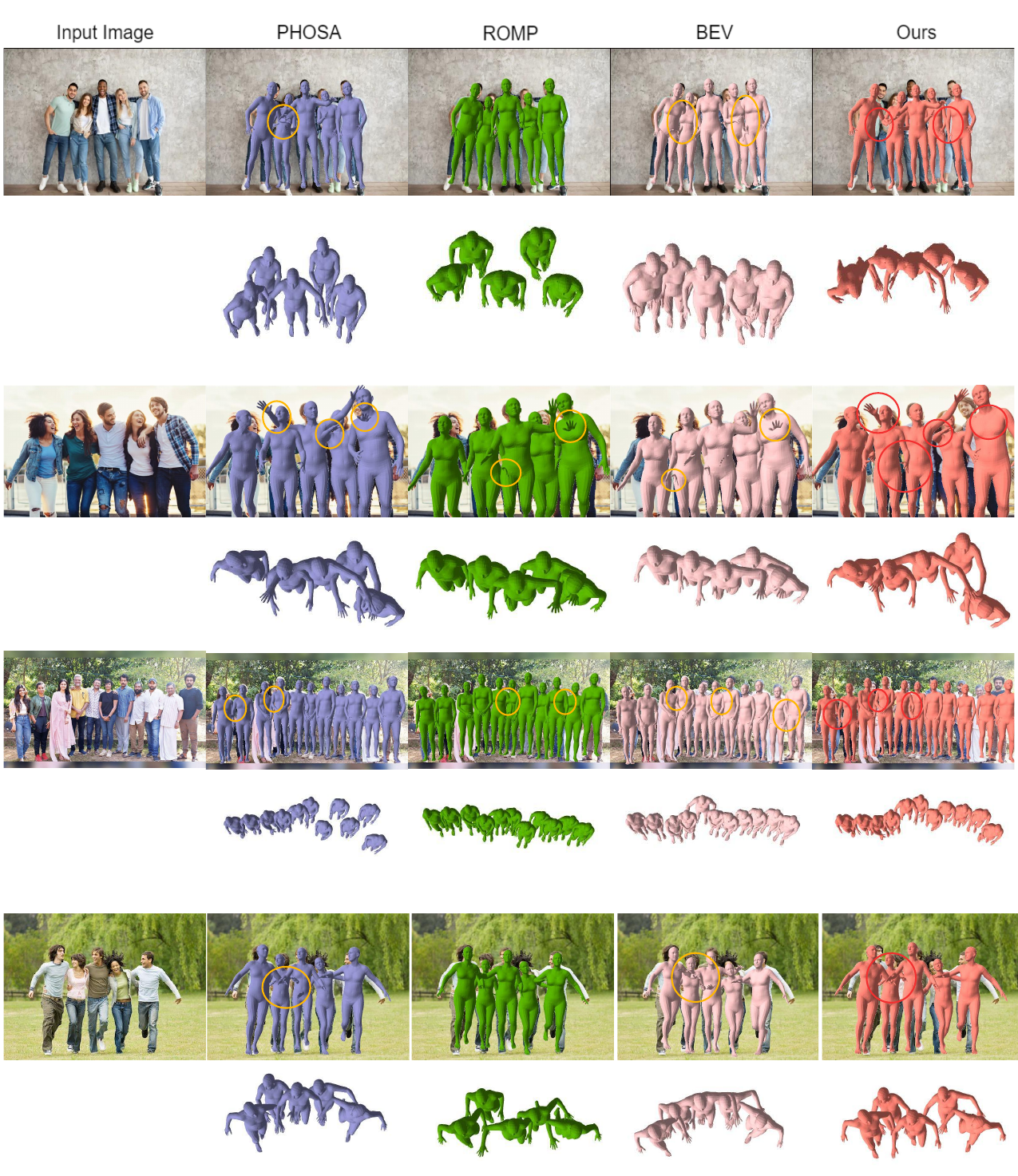}
\centering
\caption{ Qualitative results of proposed method on test images from COCO 2017 compared to PHOSA, ROMP, and BEV for human-human interactions. Our method gives more realistic and coherent reconstructions for images with multiple humans.}
\label{fig:6}
\end{figure}

Figures \ref{fig:5} and \ref{fig:6} show a \textbf{qualitative} comparison with PHOSA, ROMP and BEV. PHOSA reconstructs both humans and objects; ROMP and BEV only reconstruct humans. As seen our approach yields improved reconstruction quality by effectively mitigating ambiguities arising from mesh collisions and occlusions.

For \textbf{quantitative} evaluation, we employ a forced-choice assessment approach similar to PHOSA\cite{zhang2020perceiving} on COCO-2017 \cite{lin2014microsoft} images since there are no 3D ground truth annotations for people and objects in images in the wild. From the COCO-2017 test set, we randomly selected a sample of images and performed reconstruction on each image. We compare our method with PHOSA, ROMP, and BEV by reconstructing the scenes and comparing the degree of mesh collisions for human-human $E_{H-col}$ and human-object $E_{HO-col}$ and incorrect depth ordering for human-human $E_{H-depth}$ and human-object $E_{HO-depth}$ interactions that results from each method. This is averaged across all images to estimate values in Table \ref{table:1}. Our approach outperforms the state-of-the-art techniques for both multi-human and multi-human-object reconstruction, as well as results in a more coherent and realistic reconstruction with significantly fewer ambiguities.

\begin{table}
\centering
\resizebox{\columnwidth}{!}{%
\begin{tabular}{|p{2.5cm}||p{1.5cm}||p{1.5cm}||p{1.5cm}||p{1.5cm}|} 
 \hline
 Methods & $E_{H-col}$ & $E_{H-depth}$ & $E_{HO-col}$ & $E_{HO-depth}$ \\ [0.5ex] 
 \hline\hline
 PHOSA& 79.42 & 86.68 & 78.21  & 68.84   \\
 \hline
 ROMP& 63.51 & 74.27 & - & -  \\
 \hline
 BEV& 35.25 & 56.17 & - & -  \\
 \hline
 Ours& \textbf{16.46} & \textbf{48.37} & \textbf{26.65}  & \textbf{33.77}  \\ [1ex] 
 \hline
\end{tabular}%
}
\caption{Quantitative evaluation with PHOSA \cite{zhang2020perceiving}, ROMP\cite{sun2021monocular}, and BEV\cite{sun2022putting}. BEV and ROMP only reconstruct humans. Equations of each evaluation parameter are given in the supplementary.}
\label{table:1}

\end{table}

\begin{table}
\centering
\begin{tabular}{|c c c c|} 
 \hline
 Ours vs. & PHOSA & ROMP & BEV \\ [0.5ex] 
 \hline\hline
  & 88\% & 80\% & 74\% \\  
 \hline
 \end{tabular}
\caption{User study that gives the average percentage of images for which our method performs better on COCO-2017. $50\%$ implies equal performance.}
\label{table:2}
\end{table}

We also perform a subjective study similar to \cite{zhang2020perceiving}, where we show the reconstructions for each image from PHOSA, ROMP, BEV, and our proposed method in a random order to the users and the users mark whether our result looks better than, equal to, or worse than the other methods. We compute the average percentage of images for which our method performs better in Table \ref{table:2}. Overall, the performance of our method is relatively better than the other methods.
\begin{table}
\centering
\resizebox{\columnwidth}{!}{%
\begin{tabular}{|c c c c c|} 
 \hline
 Ours vs. & No $L_{collision}$ & No $L_{depth}$ & No $L_{interaction}$ & No $L_{occ-sil}$  \\ [0.5ex] 
 \hline\hline
  & 83\% & 62\% & 73\% & 77\% \\  
 \hline
 \end{tabular}%
 }
\caption{In the ablation study we drop loss terms from our proposed method. The higher the percentage, the more the effect of the loss term. No $L_{collision}$ implies the exclusion of both $L_{H-collision}$ and $L_{HO-collision}$. No $L_{depth}$ involves omitting $L_{H-depth}$ and $L_{HO-depth}$. No $L_{interaction}$ means we omitted the $L_{H-interaction}$ and $L_{HO-interaction}$, and lastly No $L_{occ-sil}$ corresponds to dropping the loss term defined in eq. \ref{eq:12}}
\label{table:3}
\end{table}

\subsection{Ablation Study} \label{sec:ablation}

An ablative study was conducted to assess the significance of the loss terms in Table \ref{table:3}. The identical forced-choice test similar to PHOSA\cite{zhang2020perceiving} is conducted for the complete proposed methodology (Equation \ref{eq:method}), by omitting loss terms from the proposed method and measuring the performance. Our findings indicate that the exclusion of the collision and occlusion-aware silhouette loss has the most notable effect, with the interaction loss following closely behind. 
The collision loss prevents mesh intersection and the silhouette loss guarantees that the object poses remain consistent with their respective masks.

\subsection{Additional Results}\label{sec:4.3}

\begin{figure} 
\includegraphics[scale = 0.4]{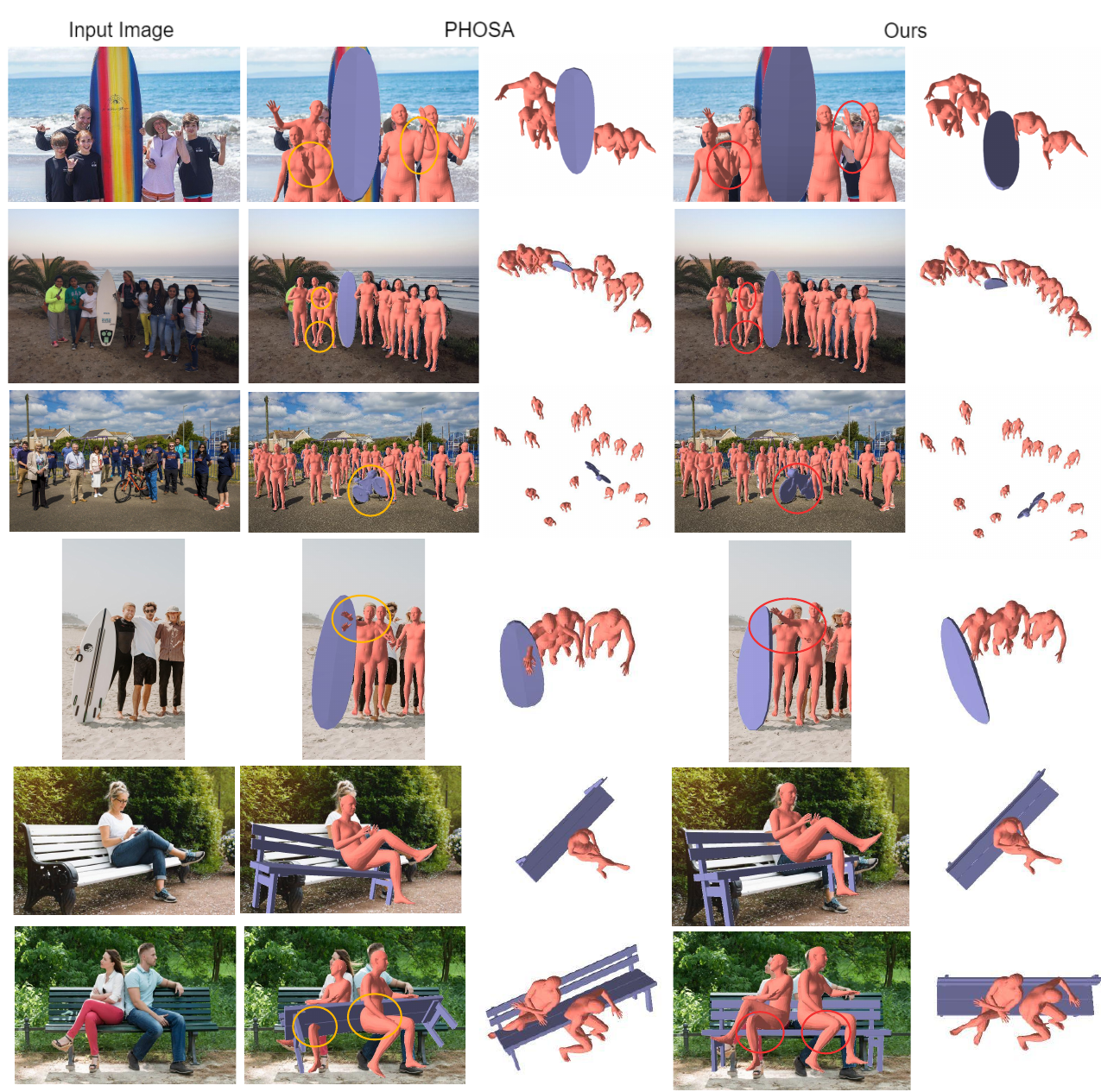}
\centering
\caption{ Our method, recovers plausible human-object and human-human spatial arrangements by explicitly reasoning about them. Here we demonstrate reconstruction on images with both humans and objects and compare PHOSA's reconstructions to those produced by our method.}
\label{fig:7}
\end{figure}

\begin{figure}
\includegraphics[scale=0.4]{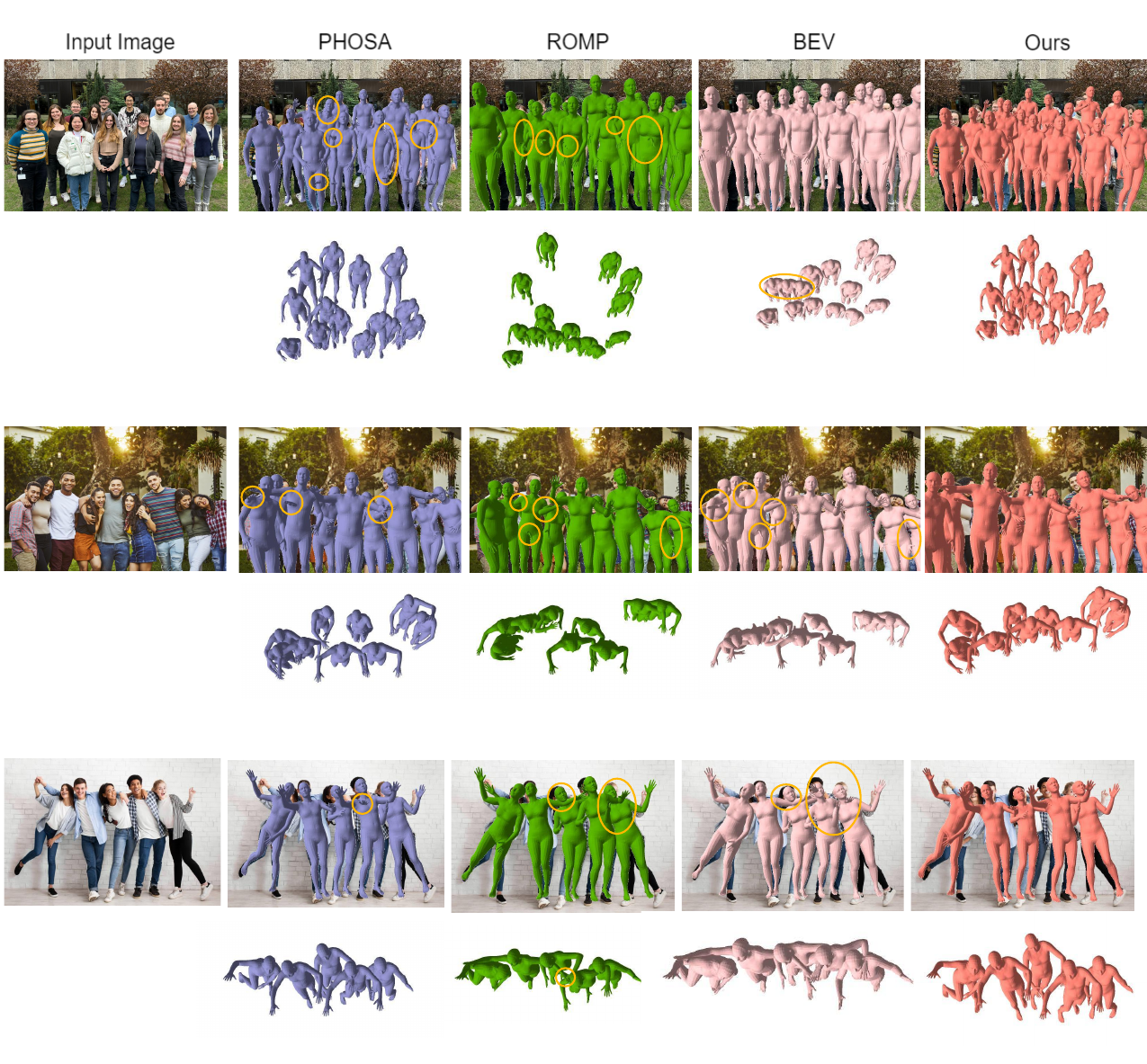}
\centering
\caption{we illustrate the differences in human reconstructions generated by PHOSA, ROMP, BEV, and Our approach when provided with an input image. Our approach produces more plausible reconstructions with a substantial decrease in mesh collisions, all while maintaining relative coherence.}
\label{fig:8}
\end{figure}

More results are shown Fig. \ref{fig:7} and \ref{fig:8} on challenging Youtube and Google images.  

\section{Discussion}

Current approaches for reconstructing humans/ objects from a single image often produce reconstructions that contain various ambiguities, especially in situations involving multiple interactions between humans and between humans and objects. 
In this paper, we perform holistic 3D scene perception by exploiting the information from both human-human and human-object interactions in an optimization framework. The optimization makes use of several constraints to provide a full scene that is globally consistent, and reduces collisions, and improves spatial arrangement (Table \ref{table:1}) over other methods.
The proposed human optimization framework resolves ambiguities between reconstructed people, and the proposed human-object optimization framework addresses ambiguities between humans and objects. 
We further introduce a method that significantly improves the pose estimation of heavily occluded objects.
We demonstrate via our qualitative and quantitative evaluations that the proposed method outperforms other methods and produces reconstructions with noticeably less ambiguity.

\section{Limitations and Future Work}
When compared to learning-based techniques, our method demands increased processing time for reconstruction generation and may occasionally yield a slightly inaccurate spatial configuration to mitigate collisions. In terms of future directions, our current implementation only considers coarse interactions among humans. However, in subsequent iterations, we aim to incorporate fine-grained interactions, leveraging this information to refine our estimation of human pose.

{
    \small
    \bibliographystyle{ieeenat_fullname}
    \bibliography{main}
}


\end{document}